\documentclass[manuscript]{acmart} 
\usepackage{graphicx, subfigure}  

\AtBeginDocument{%
  \providecommand\BibTeX{{%
    \normalfont B\kern-0.5em{\scshape i\kern-0.25em b}\kern-0.8em\TeX}}}

\begin{document}

\title{A Self-supervised Representation Learning of Sentence Structure for Authorship Attribution}



\author{Fereshteh Jafariakinabad}
\affiliation{%
 \institution{University of Central Florida}
 \streetaddress{4000 Central Florida Blvd}
 \city{Orlando}
 \state{Florida}
 \country{USA}}
\email{fereshteh.jafari@knights.ucf.edu}

\author{Kien A. Hua}
\affiliation{%
 \institution{University of Central Florida}
 \streetaddress{4000 Central Florida Blvd}
 \city{Orlando}
 \state{Florida}
 \country{USA}}
\email{kienhua@cs.ucf.edu}


\begin{abstract}
 
The syntactic structure of sentences in a document substantially informs about its authorial writing style. Sentence representation learning has been widely explored in recent years and it has been shown that it improves the generalization of different downstream tasks across many domains. Even though utilizing probing methods in several studies suggests that these learned contextual representations implicitly encode some amount of syntax, explicit syntactic information further improves the performance of deep neural models in the domain of authorship attribution. These observations have motivated us to investigate the explicit representation learning of syntactic structure of sentences. 
In this paper, we propose a self-supervised framework for learning structural representations of sentences. The self-supervised network contains two components; a lexical sub-network and a syntactic sub-network which take the sequence of words and their corresponding structural labels as the input, respectively. Due to the $n$-to-$1$ mapping of words to their structural labels, each word will be embedded into a vector representation which mainly carries structural information. We evaluate the learned structural representations of sentences using different probing tasks, and subsequently utilize them in the authorship attribution task. Our experimental results indicate that the structural embeddings significantly improve the classification tasks when concatenated with the existing pre-trained word embeddings. 
\end{abstract}

\setcopyright{acmcopyright}
\acmJournal{TKDD}
\acmYear{2020} \acmVolume{1} \acmNumber{1} \acmArticle{1} \acmMonth{1} \acmPrice{15.00}



\maketitle

\section{Introduction}

Word embeddings which can capture semantic similarities have been extensively explored in a wide spectrum of Natural Language Processing (NLP) applications in recent years.  Word2Vec \cite{mikolov2013distributed}, FastText \cite{bojanowski2017enriching}, and Glove \cite{pennington2014glove} are some examples. Even though distributional word embeddings produce high quality representations, representing longer pieces of text such as sentences and paragraphs is still an open research problem. A sentence embedding is a contextual representation of a sentence which is often created by transformation of word embeddings through a composition function. There has been a large body of work in the literature which propose different approaches to represent sentences from word embeddings. SkipThought \cite{kiros2015skip}, InferSent \cite{conneau2017supervised}, and Universal Sentence Encoder \cite{cer2018universal} are well-known examples. 

There has been a growing interest in understanding what linguistic knowledge is encoded in deep contextual representation of language. For this purpose, several probing tasks are proposed to understand what these representations are capturing \cite{tenney2019you,hewitt2019structural,conneau2018you,perone2018evaluation}. One of the interesting findings is that despite the existence of explicit syntactic annotations, these learned deep representations encode syntax to some extent \cite{blevins2018deep}. Hewitt et. al. provide an evidence that the entire syntax tree is embedded implicitly in deep model's vector geometry. Kuncoro et. al. \cite{kuncoro2018lstms} show that LSTMs trained on language modeling objectives capture syntax-sensitive dependencies. Even though deep contextual language models implicitly capture syntactic information of sentences, explicit modeling of syntactic structure of sentences has been shown to further improve the results in different NLP tasks including neural language modeling \cite {shen2017neural, havrylov2019cooperative}, machine comprehension \cite{liu2017structural}, summarization \cite{song2018structure}, text generation \cite{bao2019generating}, machine translation \cite{zhang2019syntax, li2017modeling}, authorship attribution \cite{zhang2018syntax, jafariakinabad2019style, jafariakinabad2020syntactic}, etc. Furthermore, Kuncoro et. al. provide evidence that models which have explicit syntactic information result in better performance \cite{kuncoro2018lstms}. Of particular interest, one of the areas where syntactic structure of sentences plays an important role is style-based text classification tasks, including authorship attribution. The syntactic structure of sentences captures the syntactic patterns of sentences adopted by a specific author and reveal how the author structures the sentences in a document. 

Inspired by the above observations, our initial work demonstrates that explicit syntactic information of sentences improves the performance of a recurrent neural network classifier in the domain of authorship attribution \cite{jafariakinabad2019style, jafariakinabad2020syntactic}. We continue this work in this paper by investigating if structural representation of sentences can be learned explicitly. In other words, similar to pre-trained word embeddings which mainly capture semantics, can we have pre-trained embeddings which mainly capture syntactic information of words. Such pre-trained word embeddings can be used in conjunction with semantics embeddings in different domains including authorship attribution. For this purpose, we propose a self-supervised framework using a Siamese  network \cite{chopra2005learning} to explicitly learn the structural representation of sentences. The Siamese network is comprised of two identical components; a lexical sub-network and a syntactic sub-network; which take the sequence of words in the sentence and its corresponding linearized syntax parse tree as the inputs, respectively. This model is trained based on a contrastive loss objective where each pair of vectors (lexical and structural) is close to each other in the embedding space if they belong to an identical sentence (positive pairs), and are far from each other if they belong to two different sentences (negative pairs).

As a result, each word in the sentence is embedded into a vector representation which mainly carries structural information. Due to the $n$-to-$1$ mapping of word types to structural labels, the word representation is deduced into structural representations. In other words, semantically different words (e.g. cold, hot, warm) are mapped to similar structural labels (adjective); hence, semantically different words may have similar structural representations. These pre-trained structural word representations can be used as complementary information to their pre-trained semantic embeddings (e.g. FastText and Glove). We use probing tasks proposed by Conneau et al. \cite{conneau2018you} to investigate the linguistic features learned by such a training.  The results indicate that structural embeddings show competitive results compared to the semantic embeddings, and concatenation of structural embeddings with semantic embeddings achieves further improvement.  Finally, we investigate the efficiency of the learned structural embeddings of words for the domain of authorship attribution across four datasets. Our experimental results demonstrate classification improvements when structural embeddings are concatenated with the pre-trained word embeddings.

The remainder of this paper is organized as follows: we elaborate our proposed self-supervised framework in Section \ref{method}.  The details of the datasets and experimental configuration are provided and the experimental results reported in Section \ref{results}; We review the related work in Section \ref{literature}. Finally, we conclude this paper in Section \ref{conclusion}.

\section{Related Work (Authorship Attribution)}\label{literature}

Style-based text classification is dual to topic-based text classification \cite{jafariakinabad2016maximal, oghaz2020} since the features which capture the style of a document are mainly independent of its topic \cite{argamon1998style, heidari2020using}. Writing style is a combination of consistent decisions at different levels of language production including lexical, syntactic, and structural associated to a specific author (or author groups, e.g. female authors or teenage authors) \cite{daelemans2013explanation}. Style-based text classification was introduced by Argamon-Engelson et al. \cite{argamon1998style}. The authors used basic stylistic features (the frequency of function words and part-of-speech trigrams) to classify news documents based on the corresponding publisher (newspaper or magazine) as well as text genre (editorial or news item).  Nowadays, computational stylometry has a wide range of applications in literary science \cite{kabbara2016stylistic, van2017exploring}, forensics \cite{brennan2012adversarial,afroz2012detecting,wang2017liar}, social media analysis \cite{9346546,heidari2020using,keymanesh2020twitter, 10.1145/2747880, 10.1145/3070645}, and psycholinguistics \cite{newman2003lying,pennebaker1999linguistic}. 


Syntactic n-grams are shown to achieve promising results in different stylometric tasks including author profiling \cite{posadas2015syntactic} and author verification \cite{krause2014behavioral}. In particular, Raghahvan et al. investigated the use of syntactic information by proposing a probabilistic context-free grammar for the authorship attribution purpose, and used it as a language model for classification \cite{raghavan2010authorship}. A combination of lexical and syntactic features has also been shown to enhance the model performance. Sundararajan et al. argue that, although syntax can be helpful for cross-genre authorship attribution, combining syntax and lexical information can further boost the performance for cross-topic attribution and single-domain attribution \cite{sundararajan2018represents}. Further studies which combine lexical and syntactic features include \cite{ soler2017relevance, schwartz2017effect, kreutz2018exploring}.

With recent advances in deep learning, there exists a large body of work in the literature which employs deep neural networks in the domain of authorship attribution. For instance,
Ge et al. used a feed forward neural network language model on an authorship attribution task. The output achieves promising results compared to the n-gram baseline \cite{ge2016authorship}. 
Bagnall et al. have employed a recurrent neural network with a shared recurrent state which outperforms other proposed methods in PAN 2015 task \cite{bagnall2016authorship}. 
Shrestha et al. applied CNN based on character n-gram to identify the authors of tweets. Given that each tweet is short in nature, their approach shows that a sequence of character n-grams as the result of CNN allows the architecture to capture the character-level interactions, which can then be aggregated
to learn higher-level patterns for modeling
the style \cite{shrestha2017convolutional}.  Sari et al. have proposed to use continuous representations for authorship attribution. Unlike the previous work which uses discrete representations, they represent each n-gram as a continuous vector
and learn these representations in the context of the authorship attribution tasks \cite{sari2017continuous}.

Hitchler et al. propose a CNN based on pre-trained embedding word vector concatenated with one-hot encoding of POS tags; however, they have not shown any ablation study to report the contribution of POS tags on the final performance results \cite{hitschler2017authorship}. 
Zhang et.al introduces a syntax encoding approach using convolutional neural networks which combines with a lexical models, and applies it to the domain of authorship attribution \cite{zhang2018syntax}. We propose a simpler yet more effective way of encoding syntactic information of documents for the domain of authorship attribution \cite{jafariakinabad2020syntactic}. Moreover, we employ a hierarchical neural network to capture the structural information of documents and finally introduce a neural model which incorporates all three stylistic features including lexical, syntactic and structural \cite{jafariakinabad2019style}.

\section{Proposed Framework : Lexicosynt Network} \label{method}

\begin{figure*}[h]
\centering
\includegraphics[scale=0.35]{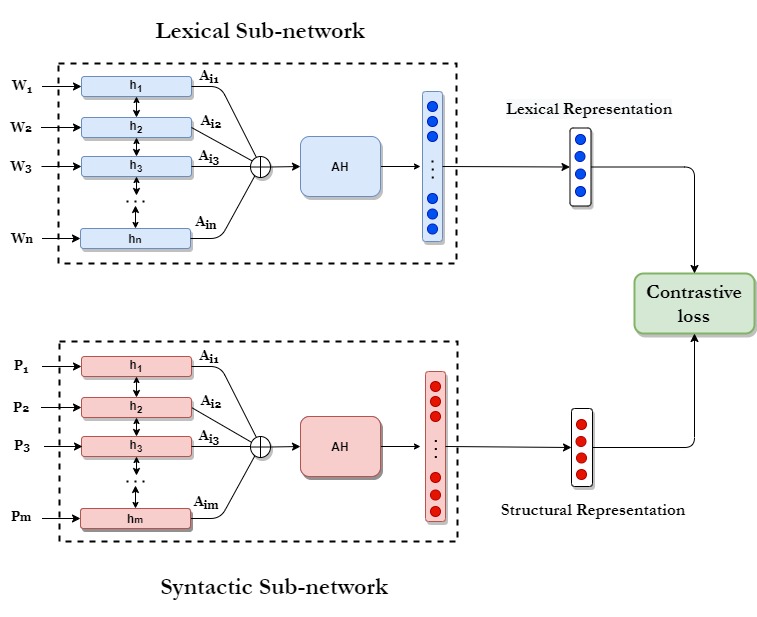}
\caption{The overall architecture of LexicoSynt network }\label{architectures}
\end{figure*}

The Lexicosynt network is a Siamese network \cite{chopra2005learning} which comprises lexical and syntactic sub-networks and a loss function. Figure \ref{architectures} illustrates the overall architecture of framework. The input to the system is a pair of sequences (lexical and syntactic) and a label. The lexical and syntactic sequences are passed through the lexical and syntactic sub-networks, respectively, yielding two outputs which are passed to the cost module. In what follows, we elaborate each component.

\subsection{Lexical Sub-network}

The lexical sub-network encodes the sequence of words in a sentence (Figure \ref{inputsequences} (a)). Each word in the sentence (denoted as $S$) is embedded into a trainable vector representation ($W_i$) and is fed into the lexical sub-network. This network consists of two main parts; bidirectional LSTM ($H$ which is concatenation of forward LSTM $\overrightarrow{h_t}$ and backward LSTM $\overleftarrow{h_t}$) and a self-attention mechanism ($A$) proposed by Lin et al. \cite{lin2017structured}. The self-attention mechanism provides a set of summation weight vectors ($W_{s2}$,$W_{s1}$) which are dotted with the LSTM hidden states, resulting weighted hidden states ($M$). Finally, a pooling layer followed by a multilayer perceptron is used to generate the final vector representation.

$$ S = (W_1, W_2, ... , W_n),$$
$$\overrightarrow{h_t} =  \overrightarrow{LSTM} (W_t, \overrightarrow{h_{t-1}} ),$$
$$\overleftarrow{h_t} =  \overleftarrow{LSTM} (W_t, \overleftarrow{h_{t+1}} ),$$
$$ H = (h_1, h_2, ... , h_n),$$
$$ A = softmax(W_{s2}tanh(W_{s1}H^T)),$$
$$ M=AH. $$

\subsection{Syntactic Sub-network}

\begin{figure}[h]
\centering
\includegraphics[scale=0.4]{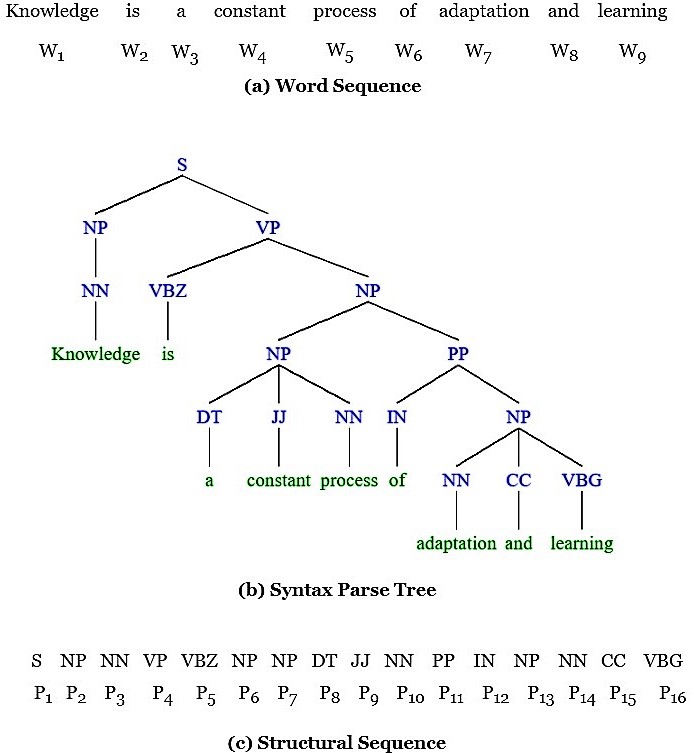}
\caption{An example of an input sentence and the corresponding (a) word sequence representation  (b) syntax parse tree representation (c) linearized parse tree representation known as structural sequence }\label{inputsequences}
\end{figure}

The syntactic sub-network aims to encode the syntactic information of sentences. For this purpose, we use syntax parse trees. Syntax parse trees represent the syntactic structure of a given sentence. An example of such a syntax tree is given in Figure \ref{inputsequences} (b). To adapt the tree representation to recurrent neural networks, we linearize the syntax parse tree to a sequence of structural labels. Figure \ref{inputsequences} (c) shows the structural label sequence of Figure \ref{inputsequences} (b) following a depth-first traversal order. Each structural label in the sequence is embedded into a trainable vector representation and subsequently is fed into the syntactic sub-network which has an identical architecture as the lexical sub-network. Needless to say the structural sequence which is the linearized syntax parse tree is longer than the word sequence of a sentence.

\subsection{Loss Function}
Given a sentence, we aim to minimize the distance of two learned vector representations from the lexical and syntactic sub-networks. The distance of two vectors ($d_n$) are maximized if they are not representations of an identical sentence. 
In other words, the output of the lexical and the syntactic sub-networks are similar ($y_n =1$) for genuine pairs (positive samples), and different ($y_n = 0$) for false pairs (negative samples). We propose to use the contrastive loss, originally proposed for training of Siamese networks \cite{chopra2005learning}, as the following:

$$E = \frac{1}{2N} \sum_{n=1}^{N} y_n d^2_n+ (1-y_n) max(margin-d_n,0)^2$$
 
where :
    $$d_n = ||V_{lexical} -V_{structural}||_2$$

$V_{lexical}$ and $V_{structural}$ are the learned sentence representations from lexical and syntactic sub-networks, respectively, and $d_n$ is the Euclidean distance of $V_{lexical}$ and $V_{structural}$. $y\in[0, 1]$ is the binary similarity metric between the input pairs; $y$ is $1$ if the syntactic and lexical pairs belong to an identical sentence and $0$ if they belong to different sentences. Parameter margin is the minimum distance between negative pairs and is set to $1$ in our implementation. Hence, the negative samples only contribute to the loss if their distance is less than the margin.

Training the network by such objective function generates similar vector representations for sentences with identical syntactic structures but different semantics. In other words, the contrastive loss objective pushes the sentences with identical syntactic structures close to each other in the embedding space. As a result, the learned vector representations from the lexical network does not mainly carry semantic relationships any more, but more structural information. Finally, the learned structural representations of words from this self-supervised framework can be simply used as complementary information to the existing ordinary pre-trained word embeddings to better suit the neural models for the domain of authorship attribution.

\section{Experimental Studies} \label{results}
In order to evaluate the effectiveness of our proposed framework, we conduct two sets of evaluations: intrinsic evaluation and extrinsic evaluation. In the former, we investigate the different linguistic properties captured by the learned structural representations and compare them against the existing pre-trained word embeddings. In the latter, we utilize the learned structural representations for the domain of authorship attribution and compare it against the existing baselines. In what follows, we elaborate the implementation details and the experimental configurations. To make the experiments reported in this paper reproducible, we have made our framework implementations publicly available \footnote{https://github.com/fereshtehjafarii/StructuralSentenceRepresentation}.

\subsection{Data}

\subsubsection{Training Data:}
The proposed model has been trained on LAMBADA (LAnguage Modeling Broadened to Account for Discourse Aspects) dataset \cite{paperno2016lambada} which contains the full text of 2,662 unpublished novels from 16 different genres. The fact that the training data comes from the wide range of genres maximizes the potential efficacy for learning diverse sentence structures. The total number of sentences in the training set is 14,746,838 where 1000 sentences are randomly selected for the development set.

\subsubsection{Test Data:}
We evaluate the proposed approach on the following authorship attribution benchmark datasets:
\begin{itemize}
\item \textbf{CCAT10 , CCAT50:} Newswire stories from Reuters Corpus Volume 1 (RCV1) written by 10 and 50 authors respectively \cite{stamatatos2008author}.

\item \textbf{BLOGS10, BLOGS50:} Posts written by 10 and 50 top bloggers respectively, originated from data set of 681,288 blog posts by 19,320 bloggers for blogger.com \cite{schler2006effects}.
\end{itemize}

\subsection{Training}

For the input data of the syntactic sub-network, we have generated the parse tree of each sentence in the training set using CoreNLP parser \cite{manning2014stanford}. Each sentence and its corresponding linearized parse tree is fed into the network as a genuine (or positive) pair. To generate the false (negative) pairs, we have paired each sentence in the batch with a randomly selected linearized parse tree in the same batch. Hence, the number of training samples is twice as the number of sentences in the training set $(14,746,838 \times 2 = 29,493,676)$. For the validation set $1000$ pairs are chosen randomly.

\begin{figure*}[h!]
\centering
\subfigure{\includegraphics[scale=0.235]{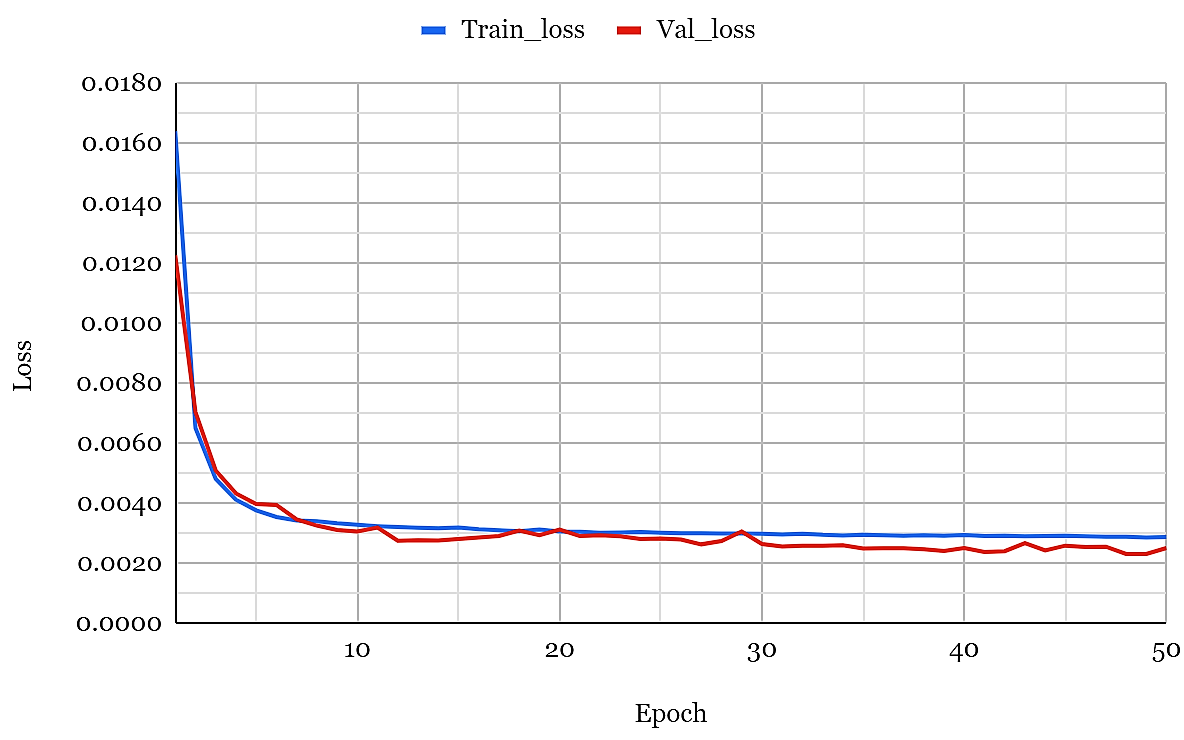}}
\subfigure{\includegraphics[scale=0.235]{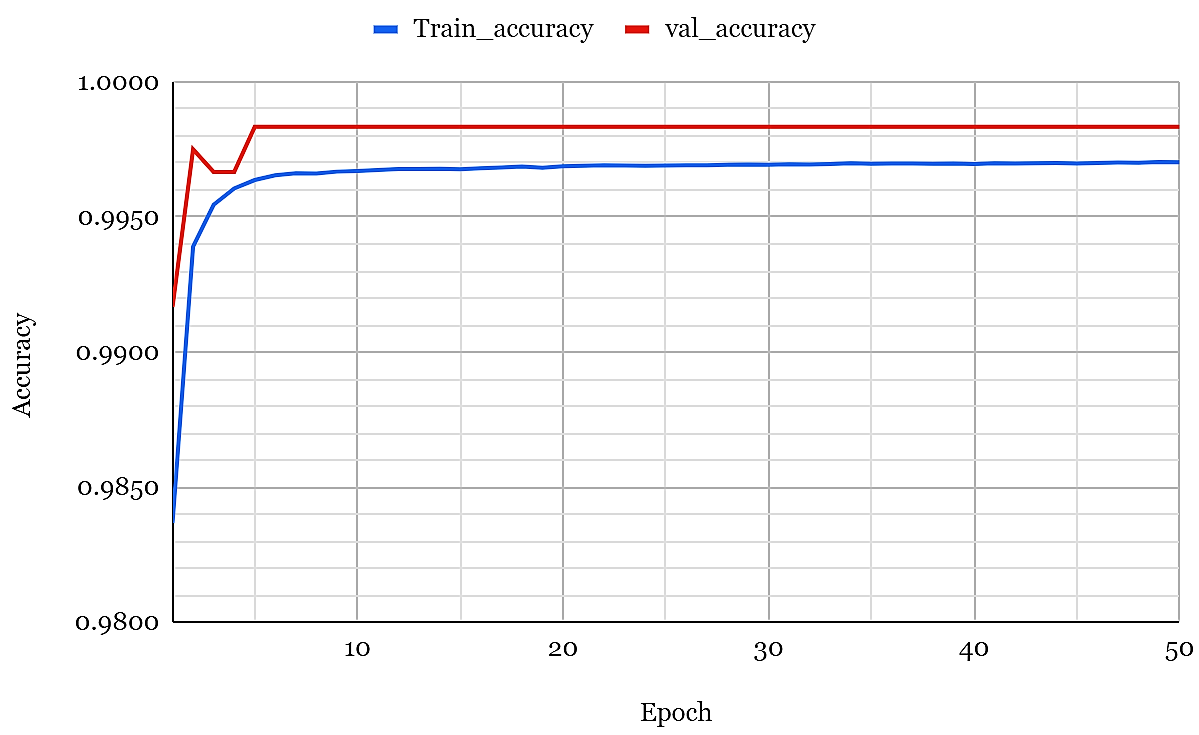}}

\caption{The training and validation loss and accuracy over 50 epochs of training }\label{train_loss}
\end{figure*}

We have used batch size of 400 and learning rate of 5e-4 with Adam optimizer for all the experiments. All the weights in the neural networks are initialised using uniform Xavier initialization. The both lexical and structural embeddings are initialized from $ U[-0.1, 0.1]$ and their dimension has been set to 300 and 100 respectively. We limit the maximum input length for lexical and syntactic sub-networks to 40 and 80 respectively. The inventory of structural labels include 77 phrase labels.
Figure \ref{train_loss} illustrates the loss and accuracy of the model across 50 epochs of training respectively. As shown in the figures, the training loss converges to 0.0029, the training accuracy is 99.70\%, and the validation loss and accuracy are 0.0023 and 99.83\% respectively. The low value of loss indicates that positive samples are successfully encoded to similar representations and negative samples are encoded to different representations.
The model performs better on the validation set compared to the training set primarily due to the fact that the training set is significantly larger ($\sim 29,000$ times) than the validation set. This is inevitable in our training scenario since training contrastive loss objective requires a huge amount of data in order to learn the proper representations. 


\subsection{Representation Learning Evaluation: Probing Tasks}

We use 10 probing tasks introduced by Conneau et al. \cite{conneau2018you} to investigate the linguistic features that the learned structural representations capture. We use the structural embeddings of words to create the Bag of Vectors (BoV) representation for each sentence and we evaluate these sentence embeddings in each task. The experiments are configured according to the recommended settings using logistic regression \footnote{https://github.com/facebookresearch/SentEval}. The probing tasks are grouped into three classes including surface information, syntactic information, and semantic information. The tasks in the surface information group do not require any linguistic knowledge while the tasks in the syntactic information group test if the sentence embeddings are sensitive to the syntactic properties. The tasks in the semantic information group not only rely on syntactic information but also require understanding of semantics about sentences. In what follows we elaborate each probing task and report the corresponding evaluation results.

\subsubsection{Surface Information} 

\begin{itemize}
    \item{\it SentLen}: To predict the length of sentences in terms of number of word.
    \item {\it WC}: To recover the information about the original word from its embedding. 
\end{itemize}

\begin{figure}[h]
\centering
\includegraphics[scale=0.23]{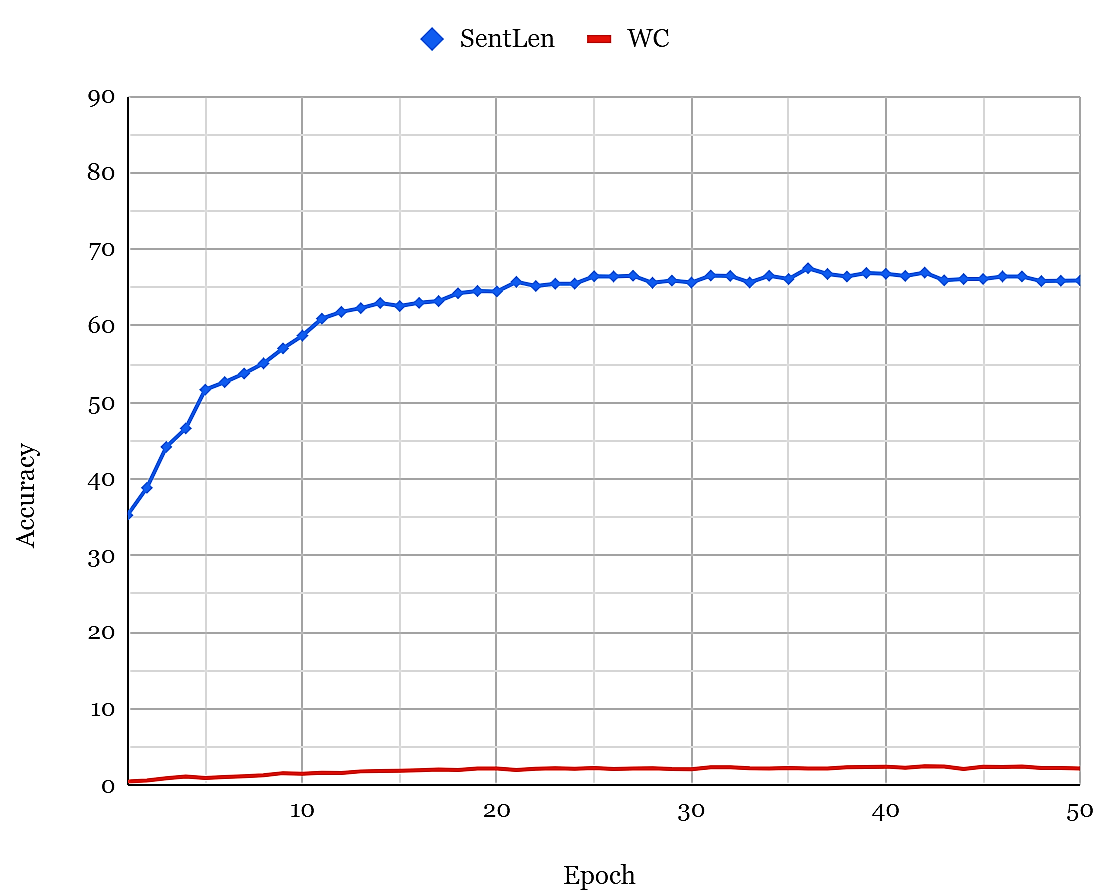}
\caption{The accuracy of learned surface information over 50 epochs of training }\label{surface_info}
\end{figure}

Figure \ref{surface_info} illustrates how surface information accuracy changes in the function of training epochs. According to the figure, the sentence length accuracy increases with epochs. Unsurprisingly, the WC accuracy is mostly flat and about 2.42\% since the model encodes structural information rather than semantic information. In other words in this model all the words with an identical syntactic role in the sentence (for instance Nouns) are mapped to a single identical vector. This way of encoding is an $n$-to-$1$ mapping ; hence, recovering the original word from its structural embedding is almost impossible. 
However, concatenating the structural embedding of words with their general embeddings (BoV structural+FastText and BoV structural+Glove) enhances the performance of WC compared to when only their general embeddings are used (Table \ref{tab:repResults}). This implies that structural information of words in the sentence improves the recovery of its content.

\subsubsection{Syntactic information}

\begin{itemize}
    \item {\it BShift}: To predict if the two adjacent words in the sentences were inverted. This task tests if the encoder is sensitive to the word order.
    \item {\it TreeDepth}: To classify sentences based on the depth of the longest path from the root to any leaf. This task investigates if the encoder infers the hierarchical structure of sentences.
    \item {\it TopConst}: To classify the sentences based on the sequence of top constituencies immediately below the sentence. This Task tests the ability of encoder in capturing the latent syntactic structure.
\end{itemize}

\begin{figure}[h]
\centering
\includegraphics[scale=0.23]{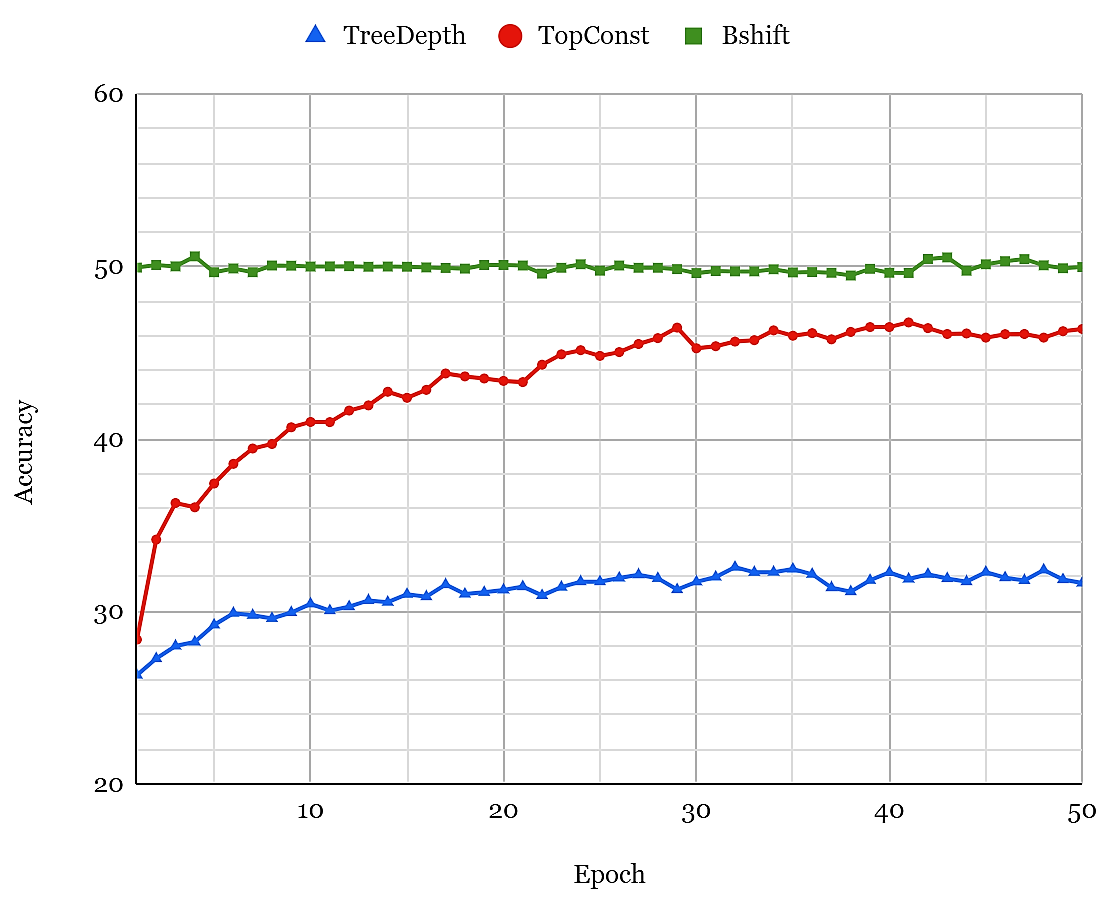}
\caption{The accuracy of syntactic information over 50 epochs of training }\label{syntactic_info}
\end{figure}

Figure \ref{syntactic_info} illustrates how syntactic information accuracy changes in terms of training epochs. According to the figure, TreeDepth and TopConst performance keep increasing with epochs. However, BShift curve is mostly flat, suggesting that what Bidirectional LSTM is able to capture about this task is already encoded in its architecture and further training does not help much.

\subsubsection{Semantic information}

\begin{itemize}
    \item {\it Tense}: To predict the tense of the main-clause verb. This task tests if the encoder captures the structural information about the main clause.
    \item {\it SubjNum}: To predict the number of the subject of main clause.  This task tests if the encoder captures the structural information about the main clause and its arguments.
    \item {\it ObjNum}: To predict the number of direct object of the main clause. This task tests if the encoder captures the structural information about the main clause and its arguments.
    \item {\it SOMO}: To predict whether an arbitrary chosen noun or verb in the sentences has been modified or not.  This task tests if the encoder has captured semantic information to some extent.
    \item {\it CoordInv}: To predict whether the order of clauses in the sentences are intact or modified. This task tests if the encoder has the understanding of broad discourse and pragmatic factors.
\end{itemize}

\begin{figure}[h]
\centering
\includegraphics[scale=0.23]{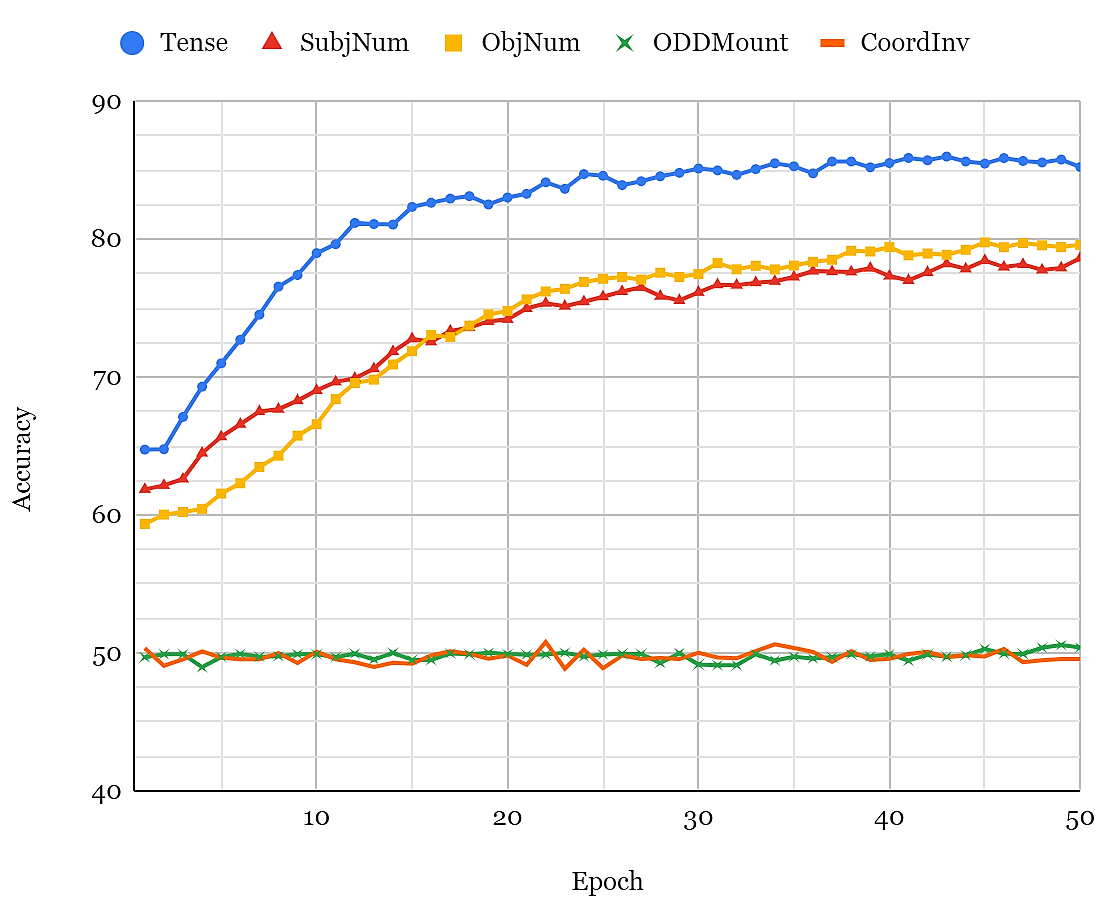}
\caption{The accuracy of semantic information over 50 epochs of training }\label{semantic_info}
\end{figure}

Figure \ref{semantic_info} illustrates how the accuracy of semantic information captured by the model changes in function of training epochs. According to the figure, the accuracy of Tense, SubjNum, and ObjNum increases when the number of epochs increase. It is worth mentioning that the accuracy of these probing tasks, which heavily rely on structural information of sentences, show more increase during the training process. On the other hand, SOMO and CoordInV which mostly rely on semantic information have flat curves, indicating that the further training does not improve their performance. This is clearly due to the fact that these two tasks deeply rely on semantic information of sentences while structural embeddings lack such information.

An interesting observation is that the structural representations generally demonstrate a better performance in the tasks where both semantic and structural information is required (e.g. Tense, SubjNum, and ObjNum) compared to the tasks that either only rely on syntactic information(e.g. TreeDepth, TopConst ) or semantic information (e.g. SOMO, CoordInv). This feature can be due to the co-supervision of lexical and syntactic sub-networks in the representation learning process.

\subsection{Representation Learning Evaluation: Comparing to the Baselines}

In this section, we compare the performance of structural embeddings learned from our model to two other pre-trained general word embeddings, including Glove \cite{pennington2014glove} and FastText \cite{bojanowski2017enriching}. We use BoV representation of sentences for both its simplicity and its capability at capturing sentence level properties \cite{conneau2018you}. Table \ref{tab:repResults} reports the results of different sentence embeddings for all the 10 probing tasks and the best results are highlighted in bold. Human. Eval. results report the human validated upper bounds for all the tasks (refer to \cite{conneau2018you} for more details). In BiLSTM Structural, we have used the syntactic sub-network as the encoder. Using BiLSTM encoder to generate the structural sentence embeddings does not show any improvement in terms of accuracy when compared to BoV structural representations except for BShift (0.07\% increase) and CoordInv (2.65\% increase) simply due to the fact that these tasks are heavily sensitive to the word orders in the sentence and BiLSTM preserves orders in the input sequence while BoV does not.

The performance results of word embeddings alone show that FastText outperforms Glove in all the tasks by the average of 1.26\% . This indicates that FastText embeddings capture slightly more linguistic features compared to Glove. Moreover, concatenating Glove/FastText embeddings with the structural embeddings (BoV Structural+Glove / BoV Structural+FastText) improves the the performance in all the tasks by the average of 2.37\% / 2.39\%. Hence, concatenating structural embeddings with the pre-trained word embeddings further improves the linguistic features captured compared to when the word embeddings are used alone.

According to the results, BoV representation of sentences from structural embeddings outperforms FastText embeddings in SentLen and TreeDepth by 12.8\% and 1.3\%, respectively. Furthermore, combining structural embeddings and FastText embeddings enhances the accuracy in all tasks compared to when only either of them is used. For instance, it improves the accuracy of ObjNum by 2.7\%, SubjNum by 2.00\%, TopConst by 4.5\%, TreeDepth by 2.6\%, SentLen by 10.7\%  compared to when only FastText embeddings are used. Unsurprisingly, combining FastText embeddings with structural embeddings does not significantly improve accuracy in WC, SOMO, and CoordInv tasks due to the fact that these tasks are heavily reliant on semantics and structural embeddings do not result in further improvements. Finally, BoV representation of sentences by combining structural embedding and FastText embeddings consistently outperforms the baseline representations in all the tasks, with an average improvement over BoV Glove and BoV FastText of 6\% and 3.9\%, respectively.

\begin{table*}[t]
\small
\tabcolsep=0.1cm
\begin{tabular}{l|cccccccccc}

\hline
Model               & \bf SentLen & \bf WC &  \bf TreeDepth & \bf TopConst & \bf BShift & \bf Tense & \bf SubjNum & \bf ObjNum & \bf SOMO & \bf CoordInv \\ \hline \hline
\multicolumn{11}{c}{\textit{ Baseline Representations}} \\\hline \hline
Majority vote       &20.0 &0.5 &17.9 &5.0 &50.0 &50.0 &50.0 &50.0 &50.0 &50.0   \\
Hum. Eval.        &100 &100 &84.0 &84.0 &98.0 &85.0 &88.0 &86.5 &81.2 &85.0    \\
BoV Glove               &58.1	&75.5	&30.0	&49.7	&49.8	&83.8	&77.2	&76.3	&49.4	&49.9\\
BoV FastText            &53.3	&79.8	&31.0	&52.6	&50.1	&86.7	&79.2	&79.4	&50.2	&50.0 \\\hline
\multicolumn{11}{c}{\textit{ Our Proposed Structural Representations}} \\\hline \hline
BiLSTM Structural       & \bf 77.8	&0.2	&23.7	&17.2	& \bf 50.3	&57.2	&50.3	&51.8	&49.9	& \bf 52.4\\
BoV Structural	       &66.1	&2.4	&32.3	&45.9	&50.1	&85.5	&78.5	&79.8	&50.3	&49.8\\
BoV Structural+Glove    &64.0	&75.1	&31.3	&54.5	&50.0	&85.4	&80.7	&81.9	&50.1	&50.4\\
BoV Structural+FastText	&64.0	& \bf 79.9	& \bf 33.6	& \bf 57.1	&50.1	& \bf 87.3	& \bf 81.2	& \bf 82.1	& \bf 50.8	&50.1\\

\hline 

\end{tabular}

\caption{\label{tab:repResults} Probing task accuracies for different sentence representations.}

\end{table*}

\subsection{Model Selection}

We have performed an ablation study on different components of the model: the self-attention mechanism, the pooling mechanism, and the length of structural sequences. Table \ref{tab:modelselection} reports the result of our experiments. In the NoATT\_seq4040 configuration, we do not use any attention mechanism and set the length of both lexical and structural sequence to 40. In WeightedATT\_seq4040, we use the traditional attention mechanism \cite{bahdanau2014neural}. In SelfATT\_AVGPool\_seq4040 and SelfATT\_MaxPool\_seq4040, we incorporate Self-attention mechanism \cite{lin2017structured} and use average-pooling and max-pooling respectively to generate the final representations. Finally, in\\   SelfATT\_MaxPool\_seq4080, we use self-attention mechanism with max-pooling where the length of lexical and structural sequence is 40 and 80, respectively . In all of the configurations other components of the network including BiLSTM and loss function, have been kept identical. According to the results, using self-attention mechanism improves the performance in most of the tasks compared to when no attention or traditional attention mechanism is used. When using self-attention mechanism, max-pooling performs better than average-pooling in most of the tasks. We observe that increasing the length of the structural sequence to 80 slightly improves the performance. This is due to the fact that structural sequence, which is a linearized syntax parse tree, is longer than the original sentence. Ultimately, the self-attention mechanism with max-pooling, and the structural sequence of length 80 is used as the final configuration (SelfATT\_MaxPool\_seq4080).

\begin{table*}[t]
\small
\tabcolsep=0.1cm
\begin{tabular}{l|cccccccccc}
\hline
Model config           & \bf SentLen & \bf WC &  \bf TreeDepth & \bf TopConst & \bf BShift & \bf Tense & \bf SubjNum & \bf ObjNum & \bf SOMO & \bf CoordInv \\ \hline \hline
\multicolumn{11}{c}{\textit{ Model Architecture for Structural Representation Learning}} \\\hline \hline

NoATT\_seq4040                             &61.6   	&5.3	    &29.7	    &43.0	    &49.7	    &85.2	    &69.5	    &71.7	    &49.9	&49.7\\
WeightedATT\_seq4040                        &62.6	    &5.1	    &30.4	    &43.4	    &50.0	    &85.3	    &68.8	    &71.4	    &49.9	&49.0\\
SelfATT\_AVGPool\_seq4040	                &62.8	    &\bf 6.6	&31.4	    &45.7	    &49.7	    &\bf 86.2	&71.9	    &74.4	    &49.9	&49.7\\
SelfATT\_MaxPool\_seq4040	                &65.9	    &2.1	    &31.6	    &45.4	    &49.9	    &85.2	    &77.9	    &79.5	    &50.0	&49.5\\
SelfATT\_MaxPool\_seq4080             	&\bf 66.1	&2.4	    &\bf 32.3	&\bf 45.9	&\bf 50.1	&85.5	    &\bf 78.5	&\bf 79.8	&\bf 50.3 &\bf 49.8\\

\hline 

\end{tabular}

\caption{\label{tab:modelselection} Probing task accuracies for different model configurations.}

\end{table*}

\subsection{Test on Authorship Attribution Datasets}

\begin{figure}[h!]
\centering
\includegraphics[scale=0.5]{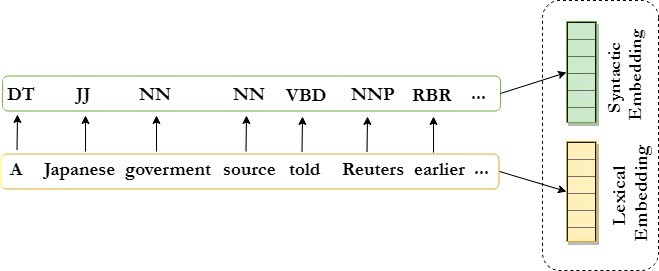}
\caption{Lexical and Syntactic Embedding }\label{jointembedding}
\end{figure}

In our previous work \cite{jafariakinabad2019style}, we have introduced a neural network which encodes the stylistic information of documents from three levels of language production (lexical, syntactic, and structural). First, we obtain both lexical and syntactic representation of words using lexical and syntactic embeddings as shown in Figure \ref{jointembedding}. 
For lexical representations, we embed each word into a pre-trained Glove embeddings and represent each sentence as the sequence of its corresponding word embeddings. For syntactic representations, we convert each word into the corresponding part-of-speech (POS) tag in the sentence, and then embed each POS tag into a low dimensional vector $P_{i} \in \mathbb{R}^{d_p}$ using a trainable lookup table $ \theta_P \in \mathbb{R}^{|T|\times d_p}$, where $T$ is the set of all possible POS tags in the language. Subsequently, these two representations are fed into two identical hierarchical neural networks which encode the lexical and syntactic patterns of documents independently and in parallel. Ultimately, these two representations are aggregated into the final vector representation of the document which is fed into a softmax layer to compute the probability distribution over class labels.

We have compared our proposed style-aware neural model (Style-HAN) with the other stylometric models in the literature, including Continuous N-gram representation \cite{sari2017continuous}, N-gram CNN \cite{shrestha2017convolutional}, and syntax-CNN \cite{zhang2018syntax}. Table \ref{ablationstudy} reports the accuracy of the models on the four benchmark datasets. All the results are obtained from the corresponding papers, with the dataset configuration kept identical for the sake of fair comparison.
In Syntactic-HAN \cite{jafariakinabad2019style}, only syntactic representation of documents is fed into the softmax layer to compute the final predictions. Similarly, in Lexical-HAN \cite{jafariakinabad2019style}, only lexical representation of documents is fed into the softmax classifier. The final stylometry model, Style-HAN, fuses both representations. In order to examine the efficacy of our proposed structural embeddings in this paper against the previously proposed POS-encoding (Syntactic-HAN) and style-aware neural network (Style-HAN), we adopt the same neural network architecture with two different settings; (1) using only structural embedding of the words (Structural-HAN) and  (2) using pre-trained Glove word embeddings concatenated with the structural embeddings (Structural+Lexical-HAN). We chose to use Glove (instead of FastText) for our performance study in order to have a fair comparison with the current state-of-the-art Lexical-HAN and Style-HAN methods, which used Glove embeddings as their lexical embeddings.\\

\begin{table}[h!]
\begin{center}
\begin{tabular}{c c c c c }
\hline \bf Model & \bf CCAT10 & \bf CCAT50 & \bf BLOGS10 & \bf BLOGS50 \\ \hline \hline
\multicolumn{5}{c}{\textit{ Baselines}} \\\hline 

Continuous n-gram & 74.8 & 72.6 & 61.3 & 52.8   \\
N-gram CNN & 86.8& 76.5& 63.7 & 53.0  \\

Syntax-CNN & 88.2 &81.0 &64.1 & 56.7 \\ 
Lexical-HAN     &86.0           & 79.5              &70.8               &59.8\\ \hline

\multicolumn{5}{c}{\textit{ Our Proposed Models}} \\\hline 

Syntactic-HAN   &63.1           & 41.3              &69.3               &57.8 \\ 
Syntactic+Lexical-HAN (Style-HAN)      &90.6           &82.3               &72.8               &61.2 \\

Structural-HAN  &65.4           &45.2               &70.6               &59.5 \\
Structural+Lexical-HAN & \bf 92.4 & \bf 83.2 & \bf 73.5 & \bf 61.7\\
\hline
\end{tabular}
\end{center}
\caption{\label{ablationstudy} The accuracy of different models for all datasets}
\end{table}

In Table \ref{ablationstudy}, the best performance result for each dataset is highlighted in bold. It shows that the proposed Structural+Lexical-HAN consistently outperforms all the baselines.  The average improvement over Continuous n-gram, N-gram CNN, Syntax-CNN, and Lexical-HAN is 18.9\%, 11\%, 7.2\%, and 5\%, respectively. These significant results confirm that explicit representation learning of syntactic structure of sentences improves the performance of lexical-based neural models in the task of authorship attribution.
Among the proposed models, Structural-HAN constantly outperforms Syntactic-HAN. This observation indicates the effectiveness of the learned structural representation of words in our proposed self-supervised framework. It is worth mentioning that the learned structural embeddings from our self-supervised framework not only improves the performance of the syntactic neural model but also eliminates the necessity of syntactic parsing in the sentence representation step in style-aware neural network; hence, it is computationally more efficient. 
Finally, Structural+Lexical-HAN is consistently the best among the proposed models across all datasets.

It is worth mentioning that in Syntactic-HAN, the syntactic units are part-of-speech tags which are embedded into randomly initialized vector representations. These syntactic representations are learned during the training phase of the Syntactic-HAN model on the authorship attribution datasets. However, in the LexicoSynt network, the units in the syntactic sub-network are part-of-speech tags which are sequenced based on the syntactic structure of sentences (linearized syntax parse tree). These structural units have been pre-trained on almost 29 million sentences in the LAMBADA dataset and subsequently used as the initialization for the Structural-HAN model. Hence, structural representations are pre-trained vector representations of part-of-speech tags which carry both structural and syntactic features of sentences. Combining structural and syntactic features of sentences is one of the advantages of Structural-HAN.

The explicit representation learning of sentence structure using the LexicoSynt network has improved the performance results of the previously proposed model in the authorship attribution task; however, learning the structural representations can be improved in different ways. The current design of the LexicoSynt network utilizes a fixed length for the lexical and structural sequences. Truncating and padding the sentences to fit this fixed length can affect their intended semantics and structures. An adaptive approach that can tailor this length to the specific case is desirable. Another limitation of the current design is due to its training on the dataset of a variety of novels. Even though the training dataset for the LexicoSynt network contains numerous novels from 16 different genres and ensures the learning of diverse sentence structures, this might limit its applicability to other domains. Domain-specific representation learning can address this limitation. 

\section{Conclusion}\label{conclusion}

In this paper, we have proposed a self-supervised framework for learning structural representation of sentences for the domain of authorship attribution. The result of training this self-supervised framework is pre-trained structural embeddings which capture information regarding the syntactic structure of sentences. Subsequently, these structural embeddings can be concatenated to the existing pre-trained word embeddings and create a style-aware embedding which carries both semantic and syntactic information and it is well-suited for the domain of authorship attribution. Moreover, structural embeddings eliminate the necessity of syntactic parsing for training syntactic neural networks; therefore, training a neural model using pre-trained structural embeddings is computationally more efficient. According to our experimental results on four benchmark datasets in authorship attribution, using structural embedding improves the performances of the proposed neural model.

\section{Acknowledgment}
This work was funded by Crystal Photonics Inc (CPI) under Grant Number 1063271. Any opinions, findings, and conclusion or recommendations expressed in
this materials are those of the authors and do not necessarily reflect the views of CPI.

\bibliographystyle{ACM-Reference-Format}
\bibliography{sample-base}

\appendix

\end{document}